\documentclass[runningheads]{llncs}

\usepackage{eccv}

\usepackage{eccvabbrv}

\usepackage{graphicx}
\usepackage{booktabs}

\usepackage[accsupp]{axessibility}  

\usepackage[pagebackref,breaklinks,colorlinks,citecolor=eccvblue]{hyperref}

\usepackage{orcidlink}

\usepackage{multirow}
\usepackage{pifont}
\newcommand{\joint}[1]{\textcolor{gray}{\textit{#1}}}
\newcommand{\pub}[1]{{\color{gray}\tiny{[{#1}]}}}
\newcommand{\xmark}{\ding{55}}
\usepackage{adjustbox}
\usepackage{subcaption}

\usepackage{ulem}

\begin{document}

\title{Background Adaptation with Residual Modeling for Exemplar-Free Class-Incremental Semantic Segmentation} 

\titlerunning{Background Adaptation with Residual Modeling}

\author{Anqi Zhang\orcidlink{0000-0002-2112-8199} \and
Guangyu Gao\thanks{Corresponding Author.}\orcidlink{0000-0002-0083-3016}}

\authorrunning{A.~Zhang et al.}

\institute{School of Computer Science and Technology,\\ 
Beijing Institute of Technology, Beijing, China \\
\email{andy\_zaq@outlook.com}\\
\email{guangyugao@bit.edu.cn}}

\maketitle

\begin{abstract}
Class Incremental Semantic Segmentation~(CISS), within Incremental Learning for semantic segmentation, targets segmenting new categories while reducing the \textit{catastrophic forgetting} on the old categories.
Besides, \textit{background shifting}, where the background category changes constantly in each step, is a special challenge for CISS. 
Current methods with a shared background classifier struggle to keep up with these changes, leading to decreased stability in background predictions and reduced accuracy of segmentation.
For this special challenge, we designed a novel background adaptation mechanism, which explicitly models the background residual rather than the background itself in each step, and aggregates these residuals to represent the evolving background. 
Therefore, the background adaptation mechanism ensures the stability of previous background classifiers, while enabling the model to concentrate on the easy-learned residuals from the additional channel, which enhances background discernment for better prediction of novel categories. 
To precisely optimize the background adaptation mechanism, we propose Pseudo Background Binary Cross-Entropy loss and Background Adaptation losses, which amplify the adaptation effect. 
Group Knowledge Distillation and Background Feature Distillation strategies are designed to prevent forgetting old categories.
%
Our approach, evaluated across various incremental scenarios on Pascal VOC 2012 and ADE20K datasets, outperforms prior exemplar-free state-of-the-art methods with mIoU of 3.0\% in VOC 10-1 and 2.0\% in ADE 100-5, notably enhancing the accuracy of new classes while mitigating catastrophic forgetting. 
Code is available in \href{https://andyzaq.github.io/barmsite/}{https://andyzaq.github.io/barmsite/}. 
\end{abstract}

\section{Introduction}
\label{sec:intro}


With the development of deep learning and the support of convolutional neural networks (CNNs)~\cite{lenet98,alexnet12, vggnet15,resnet16} and Transformers~\cite{vaswani2017attention,vit20,swin21,deit21}, current semantic segmentation networks~\cite{fcn,unet,deeplabv1,deeplabv2,deeplabv3,chen2018encoder,maskformer,mask2former,ln23} have achieved remarkable results on standard supervised learning tasks. 
However, real-world data, arriving as an unstable continuous stream, poses the challenge of retaining prior knowledge while accommodating new concepts, termed as \textit{catastrophic forgetting}~\cite{catastrophic89,there95}.

To tackle this problem, \textit{incremental learning} is proposed to adapt to changing data streams for new concepts, but also avoid forgetting old knowledge, especially for the classification task, \ie, Class-Incremental Learning~(CIL)~\cite{lfl16,lwf17,lifelong18,slca23,ctp23}.
Cermelli et al.~\cite{mib20} introduced CIL into Semantic Segmentation as Class Incremental Semantic Segmentation~(CISS). 
Since the semantic segmentation tasks involve dense predictions, the issue of catastrophic forgetting typically becomes more challenging.
Most recent works~\cite{mib20,ssul21,plop21} have struggled to mitigate \textit{catastrophic forgetting} through knowledge distillation~\cite{plop21, sdr21, dkd22} or parameter freezing~\cite{ssul21,micro22}. 
Although these strategies effectively alleviate \textit{catastrophic forgetting}, it comes at the cost of the model's plasticity, making it challenging to learn novel classes.
In current state-of-the-art approaches~\cite{micro22,dkd22}, the performance of established old classes approaches the upper limit of \textit{joint} training, yet there still remains a significant gap in mIoU for novel classes. 
However, in the context of \textit{lifelong learning} scenarios featuring an infinite influx of novel classes, the incremental learner's plasticity becomes increasingly crucial. 
Therefore, the optimal solution should be fine-tuning more parameters for novel classes, while \textit{catastrophic forgetting} needs to be handled properly. 
\begin{figure}
    \centering    \includegraphics[width=0.9\linewidth]{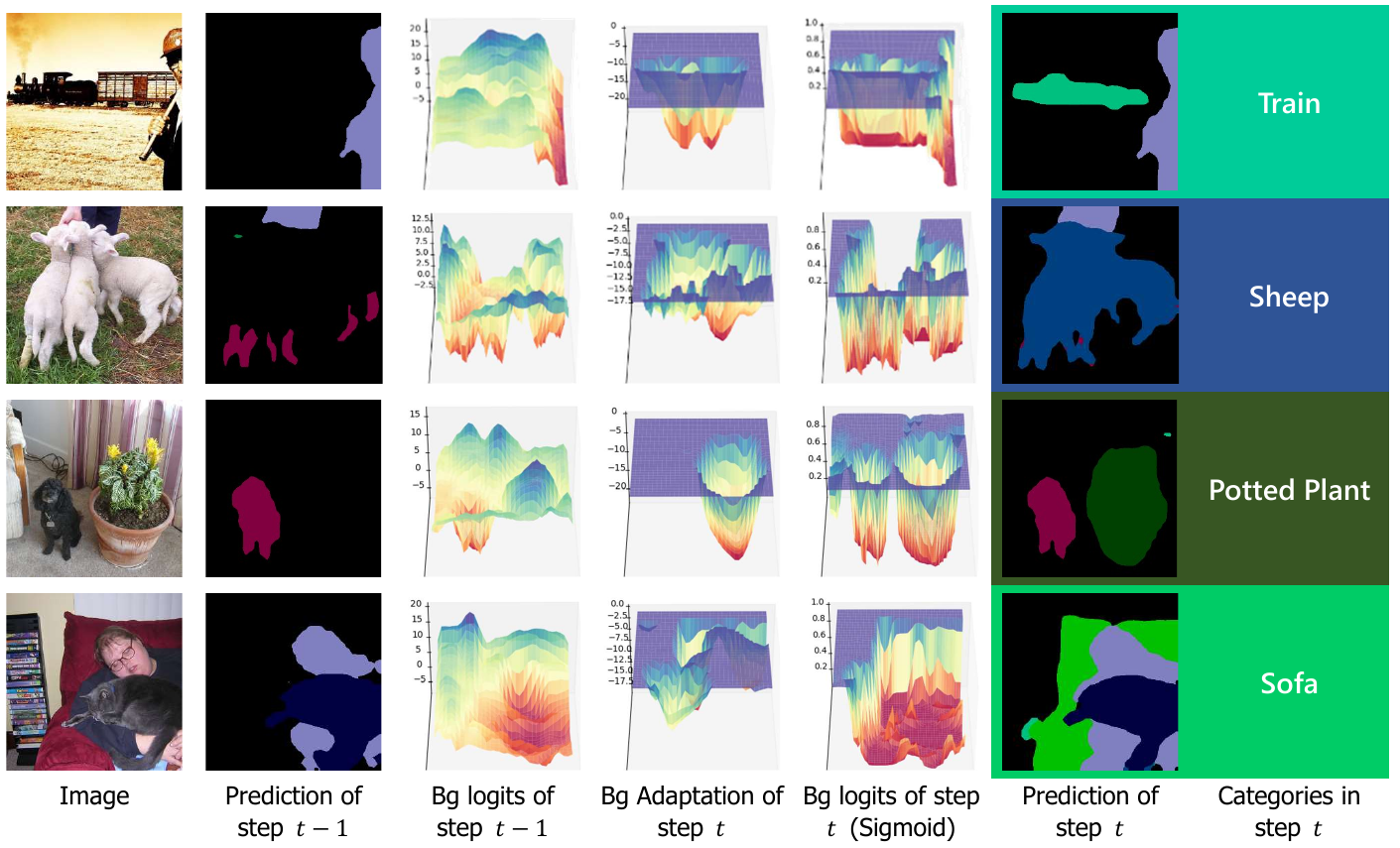}
    \caption{3D Visualization of Background Adaptation results. 
    The background logits for step $t$ combine those from step $t-1$ with the adaptation logits learned in step $t$, preventing disorderly adjustments and focusing on residuals.
    The current step background logits are processed using the Sigmoid function. 
    Note that colors range from red~(small values) to blue~(large values).}
    \label{fig:bga}
\end{figure}

Nevertheless, \textit{background shifting}, where the background category changes constantly in each step, is another challenge for CISS.
Upon further investigation, it was observed that previous methods within CISS often utilize a shared background classifier across each step. 
However, novel classes are typically considered as the background in previous learning steps, indicating that the background class undergoes continuous changes throughout incremental learning.
This way results in an incessantly shifting background in each step, leading to disruption in background prediction and inaccurate learning of novel classes.
Some approaches~\cite{ssul21,star23,alife22} store old exemplars in extra memory space to prevent \textit{background shifting}, while long-term storage of such data may not be feasible in real-world scenarios.
To enhance the \textit{incremental} nature of the CISS method and reduce background instability, we revise the representation of the background class and propose an exemplar-free background adaptation mechanism. 
In particular, the background adaptation mechanism explicitly models the background residual rather than the background itself in each incremental step. 
The learned residuals of the additional channel are aggregated with the previously learned background logits to represent the evolving background, as illustrated in Fig.~\ref{fig:bga}. 
By employing the background adaptation mechanism, we only need to optimize the additional channel of residuals for constructing the background during the incremental steps. 
This approach eliminates the need for repeatedly learning background classifiers associated with old classes, thereby avoiding the instability of optimizing a shared background classifier.

Moreover, we customize our training approaches in order to fit the training process of the background adaptation structure and promote its performance.
We first introduce the widely used pseudo-labeling~\cite{plop21, ssul21} to generate a pseudo-background, and then we design a PB-BCE loss to focus on optimizing the classifier of novel classes and related background adaptation residual. 
Besides, for the Background Adaptation channel, the objective is to minimize the values in the regions of novel categories, while ensuring that the values in other regions remain above 0.
Thus, two auxiliary Background Adaptation losses are specifically applied to the background adaptation channel for precise adjustment. 
Given our pursuit of plasticity on novel categories without external memory, we have devised more comprehensive exemplar-free knowledge distillation strategies to combat \textit{catastrophic forgetting}.
We further adopt knowledge distillation strategies on the logit maps of old categories and the intermediate features, namely the Group Knowledge Distillation and Background Feature Distillation, respectively. 
Finally, our approach outperforms previous methods in multiple benchmarks, especially in realistic and challenging long-term scenarios such as 10-1 of VOC2012 and 100-5 of ADE20K. 
Our approach achieves significant performance gains of \textbf{4.5\%} and \textbf{5.2\%} on novel classes, respectively, compared to the previous state-of-the-art.

Overall, our contributions are summarized as follows: 
\begin{itemize}
    \item We propose a Background Adaptation strategy in incremental steps by optimizing an additional channel for residual modeling instead of the shared background classifier, which avoids unstable optimization and prevents chaos in background logits. 
    \item We design PB-BCE loss and Background Adaptation losses to enhance the effectiveness of Background Adaptation, while two distillation strategies ensure maintaining knowledge of old categories. 
    \item Experiments validate the effectiveness of our proposed method, showcasing a new state-of-the-art performance on CISS benchmarks.
\end{itemize}

\section{Related Works}
\label{sec:formatting}

\subsection{Class Incremental Learning}

Class Incremental Learning~(CIL) breaks the constraints of the standard training process in an incremental form on new classes. 
CIL task aims to continually learn the new classes without retraining on the data of old classes, which significantly reduces the training cost. 
As a compromise, the CIL task has to effectively solve the problem of \textit{catastrophic forgetting}~\cite{catastrophic89, there95}. 
Current CIL methods are divided into several main categories. 
Replay-based methods recall the data of old classes during incremental training. Some of the approaches~\cite{gradient17, gss19, ccbo20} reserve memory buffers and temporarily store representative past data. 
Others~\cite{dgr17, mergan18, dgm19} introduce powerful generative models to generate pseudo-data for rehearsal. 
Regularization-based methods introduce regularization terms to prevent forgetting the old classes. 
These methods utilize loss functions as penalties~\cite{ewc17, ritter2018online, schwarz2018progress, ucl19} or knowledge distillation from the previous model~\cite{lwf17, icarl17, lwm19}. 
Representation-based methods attempt to transfer representations to certain categories, which apply adaptation on a fixed pre-trained backbone~\cite{side20, dlcft22, twf22, ada22} or self-supervised learning~\cite{lump21, co2l21, dualnet21}. 

\subsection{Class Incremental Semantic Segmentation}

The achievements in CIL tasks encourage the Incremental Learning pipeline to develop in other dense prediction tasks, including Semantic Segmentation. 
Class Incremental Semantic Segmentation~(CISS) task is first introduced in MiB~\cite{mib20}, which focused on reconstructing the background region under the restriction from knowledge distillation. 
Since then, various methods~\cite{ucd22, idec23} have been introduced in the CISS task. PLOP~\cite{plop21} defined the pseudo-labeling and multi-scale local distillation to remember the old categories. 
SDR~\cite{sdr21} introduced category-wise prototype matching and contrastive learning methods to enlarge the difference between categories. RCIL~\cite{rcil22} and EWF~\cite{ewf23} explored methods for integrating parameters of new and old models. 
DKD~\cite{dkd22} decomposed the class logits into positive and negative logits for better knowledge distillation and applied threshold to generate background regions. 
Incrementer~\cite{incre23} borrows prompt learning from ViT~\cite{vit20} and applies the structure of Segmenter~\cite{segmenter21} instead of DeeplabV3~\cite{deeplabv3}, which is the basic structure of other methods. 
Besides, several approaches utilize external memory~\cite{amss23} or models~\cite{recall21,coinseg23} for better performance. 
SSUL~\cite{ssul21} introduced a Saliency-map detector to identify the saliency region and store some of the previous instances in memory. 
MicroSeg~\cite{micro22} utilized proposals from pretrained Mask2Former~\cite{mask2former} as a guidance. 
ALIFE~\cite{alife22} memorized category features and adapted for remembering the old classes. 

\section{Methods}


\begin{figure*}[!t]
    \centering
        \subcaptionbox{The overall pipeline of our approach. \label{fig:frame}}{
		\includegraphics[width=0.76\linewidth]{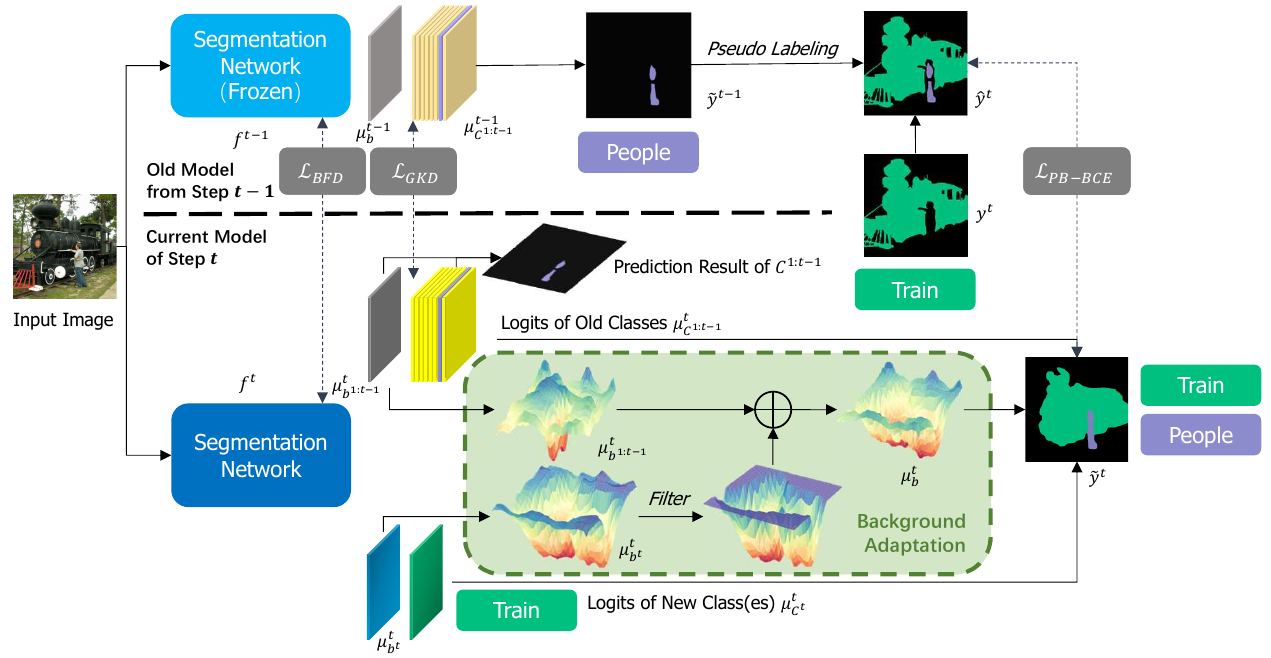}		
	}
	\hfill 
	\subcaptionbox{BgA losses. \label{fig:bgaloss}}
 {		\includegraphics[width=0.19\linewidth]{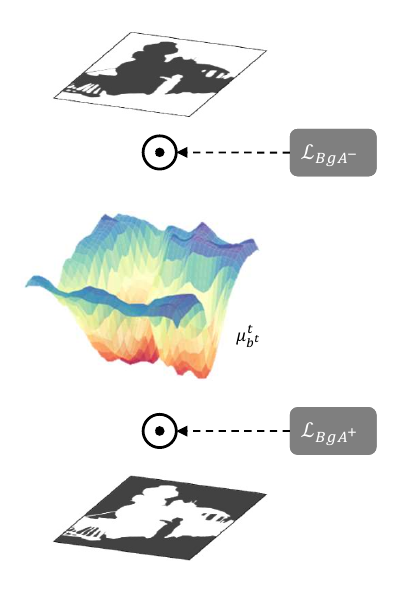}
 }
    \caption{Overview of our framework that consists of a Background Adaptation mechanism and proposed losses. 
    (a) The Background Adaptation mechanism uses the negative part of the adaptation channel to rectify the background logits. 
    (b) Two Background Adaptation losses separately optimize the regions \textit{w/} and \textit{w/o} novel categories to enhance the performance of the Background Adaptation. }
    \label{fig:overall}
\end{figure*}

\subsection{Preliminaries}

The Class-Incremental Semantic Segmentation~(CISS) task performs training in $T$ steps. 
During step $t$, the training data $D^t$ containing the labels of the novel classes $C^t$ are applied for training, while the labels of old classes $C^{1:t-1}$ are expunged. 
In particular, each sample of the training data $D^t$ consists of an RGB image $x^t \in \mathbb{R}^{3\times H\times W}$ and a corresponding ground truth label $y^t$ on novel classes $C^t$. 
Besides, the ground truth label $y^t$ replaces the labels of the $C^{1:t-1}$ with the background label $c_b$ as the old classes will not appear in $C^t$ to ensure $C^{1:t-1} \cap C^t = \varnothing$. 
Therefore, the model $f^t$ being trained at step $t$ not only needs to learn to segment novel categories but also has to retain the capability to segment old categories. 
However, while learning the novel categories, the model $f^t$ tends to overfit the novel categories $C^t$ as $f^t$ cannot contact with samples of $C^{1:t-1}$, causing severe \textit{catastrophic forgetting}. 
Moreover, the novel categories can only appear in the background region during the previous steps. 
We are more likely to classify them from the previous background prediction region, which motivates us to focus on background adaptation.

\subsection{Background  Adaptation \label{sec:bgadp}}

Previous studies often apply only one background classifier. 
They have to readjust the parameters of the background classifier whenever the model learns novel categories, which causes \textit{catastrophic forgotten} even if the classifier of the old categories remains the same probability output. 
The unstable background classifier could be a thorny constraint factor. 
To tackle the problems above, we design a Background Adaptation mechanism. 
Different from the previous distinct background classifier, we introduce an additional background adaptation channel for each classifier of the incremental process. 

During the training process of incremental step $t$, the current model $f^t$ inherits the parameters from the previous model $f^{t-1}$ and adds a new classifier $f_{C^t}^t$ for parallel classification of categories in $C^t$ alongside the previous classifiers. 
The encoder $f_e^t$ and the decoder $f_d^t$ transform the input image $x^t$ into features map $F_d^t \in \mathbb{R}^{h\times w\times c}$:
\begin{equation}
    F^t_d = f_d^t(f_e^t(x^t)). 
\end{equation}
Compared to the standard classifier of the previous methods~\cite{plop21,ssul21,alife22}, our classifier of each incremental step has an additional channel for background adaptation. 
Therefore, for each group of categories $C^i$ from step $i$, the classifier $f_{C^i}^t$ of the model $f^t$ generates the corresponding classification logit maps $\mu_{C^i \cup \{b_i\}}^t \in \mathbb{R}^{h\times w\times (|C^i|+1)}$: 
\begin{equation}
    \mu_{C^i \cup \{b_i\}}^t = f_{C^i}^t(F^t_d), 
\end{equation}
where $i\in \{1, 2, \ldots, t\}$.  
We define $b^1$ to represent the background class of the initial step, and $b^i$ denotes the background adaptation for subsequent incremental steps when $i>=2$. The symbol $b$ refers to the adapted background class. 
Thus, $\mu_{C^i \cup \{b_i\}}^t$ contains the probability maps $\mu_{C^i}$ of $|C^i|$ categories from step $i$ and an additional channel $\mu_{b^i}^t$ for constructing the background logits $\mu_b^t$. 
Considering that the region of the novel categories $C^t$ is more likely inherited from the background region of the previous step, the classifier $f^t_{C^t}$ adjusts the background logits through $\mu_{b^t}^t$, which decreases the probability values in the region $\mathcal{R}_{C^t}$ of $C^t$. 
The parameters of previous classifiers are not required to continuously optimize for new background regions, thereby avoiding instability in the classification of old categories.

The background adaptation channel $\mu_{b^t}^t$ can correct the background logits $\mu_b^t \in \mathbb{R}^{h\times w\times 1}$ in $\mathcal{R}_{C^t}$, yet the noise in other regions of $\mu_{b^t}^t$ may interfere with the background prediction. 
Particularly, when old categories are present in the image, there is a risk that these regions could be incorrectly covered by the adapted background.
Hence, we design a \textit{Filter} operation that replaces positive values of $\mu_{b^i}^t$ with $0$. 
As shown in Fig.~\ref{fig:frame}, with the \textit{Filter} operation applied, the residual modeling of background logits during the inference process is as follows: 
\begin{equation}\label{func:adp}
    \mu_b^t = \mu_{b^1}^t + \sum_{i=2}^{t} Filter(\mu_{b^i}^t), 
\end{equation}
where the background logits from the initial step $\mu_{b^1}^t$ stay the same and the background adaptation logits from the incremental steps $\mu_{b^i}^t$ only remain the negative values via the \textit{Filter}. 
In particular, to make $\mu_b^t$ more friendly for gradient backpropagation, we abandoned the \textit{Filter} on the background adaptation logits of novel classes $\mu_{b^t}^t$ during the training process: 
\begin{equation}
\label{func:adptrain}
    \mu_b^t = \mu_{b^1}^t + \sum_{i=2}^{t-1} Filter(\mu_{b^i}^t) + \mu_{b^t}^t. 
\end{equation}

\subsection{Incremental Training Strategy}

\subsubsection{Pseudo Labeling.}\label{sec:pbbce}

During the incremental steps, even if objects of old categories $C^{1:t-1}$ exist in the image, we cannot obtain corresponding labels from the ground truth. 
In order to simulate a ground truth label that contains the categories we have learned, we follow~\cite{plop21,ssul21,micro22} to introduce the prediction probability of the previous step ${\varphi}^{t-1} \in \mathbb{R}^{ h \times w \times |C^{1:t-1}|}$ from $f^{t-1}(x^t)$ with Sigmoid function and combine it with the current ground truth label $y^t$:
\begin{equation}
    \hat{y}^{t} = 
    \begin{cases}
        y^t, & (y^t \in C^t)\vee \{ (y^t \in c_b) \wedge (max({\varphi}^{t-1}) < \tau)\} \\
        \tilde{y}^{t-1}, & (y^t \in c_b) \wedge (max({\varphi}^{t-1}) \geq \tau)
    \end{cases}
\end{equation}
where $\hat{y}^t \in \mathbb{R}^{h\times w}$ denotes the generated pseudo label and $\tilde{y}^{t-1} = argmax({\varphi}^{t-1})$ represents the final prediction from $f^{t-1}(x^t)$. 

\subsubsection{PB-BCE Loss.}

Binary Cross-Entropy~(BCE) loss is extensively employed for learning to segment objects of novel categories. 
We follow the usage of the Sigmoid function and activate $\mu^{t}$ as ${\varphi}^t$. 
Some of the CISS methods~\cite{plop21, ssul21, micro22} utilize the pseudo mask as the ground truth for supervision, which effectively replaces the background label in the regions of the old categories. 
However, the ambiguous supervision of the old categories could improperly optimize the classifiers of $C^{1:t-1}$ and destroy their identification capability. 
Therefore, we design a Pseudo Background BCE~(PB-BCE) loss which cuts off the backpropagation of old categories, meaning that we only take the labels of current categories $\hat{y}^t_{C^t}$ and pseudo-background $\hat{y}^t_{c_b}$ for optimization: 
\begin{equation}
    \mathcal{L}_{PB-BCE} = -\frac{1}{hw}\sum_{k=1}^{hw}\sum_{c\in\{c_b\}\cap C^t} CE({\varphi}^t_{c,k}, \hat{\psi}^t_{c,k}), 
\end{equation}
where $k$ represents the index of pixel and $\hat{\psi}^t_c$ denotes the binary map of class c in pseudo map $\hat{y}^{t}$. $CE$ is the cross-entropy calculation:
\begin{equation}
    CE({\varphi}^t_*, \hat{\psi}^t_*) = \hat{\psi}^t_*\cdot log {\varphi}^t_* + (1 - \hat{\psi}^t_*)\cdot log (1 - {\varphi}^t_*). 
\end{equation}
Although the whole background prediction logits $\mu_b^t$ are involved in the loss calculation, only the background adaptation channel for current categories $\mu_{b^t}^t$ serves for optimization. Previous background logits are detached to prevent disorderly evolving on parameters of the old classifiers. 

\subsubsection{Background Adaptation Losses.}
\label{sec:bgaloss}

The Background Adaptation~(BgA) losses, as shown in Fig.~\ref{fig:bgaloss}, are designed for further refined adjustments on the background adaptation channel $\mu_{b^t}^t$. 
We propose two losses for two different regions of $\mu_{b^t}^t$ according to the target effect. 
For regions $\mathcal{R}_{C^t}$ belonging to the current classes $C^t$, the ideal target is to have negative values as small as possible. 
To achieve this, we introduce cross-entropy loss $\mathcal{L}_{BgA^+}$ for supervision: 
\begin{equation}
    \mathcal{L}_{BgA^+} = -\frac{1}{|\mathcal{R}_{C^t}|}\sum_{k \in \mathcal{R}_{C^t}}log\frac{exp(-\mu_{b^t,k}^t)}{1+exp(-\mu_{b^t,k}^t)}, 
\end{equation}
where $|\mathcal{R}_{C^t}|$ represents the number of pixels in $\mathcal{R}_{C^t}$. 
Besides, as for the regions of the background that might contain old classes, we hope the value of $\mu_{b^t}^t$ could be 0 for maintaining the historical background prediction result. 
However, achieving precise control over the whole region to be a precise value is challenging and may be impractical. 
Therefore, we set the positive logits in the adaptation channel of the old categories to 0 in Eq.~\ref{func:adptrain} to avoid the noise and apply a Triplet loss $\mathcal{L}_{BgA^-}$ to supervise the background region: 
\begin{equation}
    \begin{aligned}
        \mathcal{L}_{BgA^-} = \frac{1}{hw-|\mathcal{R}_{C^t}|}\sum_{k\notin \mathcal{R}_{C^t}} max[0, (1-\varphi_{b^t,k}^t)^2 - (\varphi_{b^t,k}^t-0)^2]. 
    \end{aligned}
\end{equation}
We first activate the background adaptation channel as $\varphi_{b^t}^t$ via Sigmoid function, then the logits are mapped to the range from 0 to 1. 
We aim for the logits $\mu_{b^t}^t$ of this region to be greater than 0 while not overfitting to the objects in the background, meaning that values of $\varphi_{b^t}^t$ lie within the range of 0.5 to 1 and not close to 1. 
In particular, The Triplet loss $\mathcal{L}_{BgA^-}$ respectively set 0 and 1 as negative and positive anchors. 
We simultaneously constrain the loss not to go below 0 by setting a lower bound. 
Therefore, once the value of the pixel exceeds 0.5, \ie, the middle of the positive and negative anchors, there will be no more gradient on the pixel for backpropagation, which mitigates overfitting to the objects in the background.

\subsubsection{Knowledge Distillation Strategies.}\label{sec:kd}

In the training process of incremental steps for CISS tasks, substantial alterations to the existing parameters of $f^t$ frequently occur, giving rise to situations of \textit{catastrophic forgetting}. 
Guided by the previous methods, we have devised a series of knowledge distillation strategies tailored to our background adaptation mechanism. 
\begin{itemize}
    \item 
\textbf{Group Knowledge Distillation.} In Sec. \ref{sec:bgadp}, we have introduced the classification logit map $\mu_{C^i \cup \{b_i\}}^t$ for each group of categories $C^i$, which contains an additional channel for background adaptation. 
We introduce a Group Knowledge Distillation loss $\mathcal{L}_{GKD}$ to ensure the stability of the old category logits, as well as the previous background adaptation channel: 
\begin{equation}
    \mathcal{L}_{GKD} = -\frac{1}{hw}\sum_{k=1}^{hw} \sum_{i=1}^{t-1} \sum_{c \in \{c_b^i\}\cup C^i} 
    CE(\varphi_{c,k}^{t}, \varphi_{c,k}^{t-1})
\end{equation}
where $\varphi_*^{t-1}$ and $\varphi_*^{t}$ denote the activated logits of $\mu_*^{t-1}$ and $\mu_*^{t}$, respectively. 
\item
\textbf{Background Feature Distillation.} 
When the network learning novel categories, not only the parameters of the novel classifier $f^t_{C^t}$ are optimized, but the encoder $f^t_e$ and the decoder $f^t_d$ are affected as well. 
Merely utilizing Group Knowledge Distillation loss $\mathcal{L}_{GKD}$ is insufficient in effectively mitigating \textit{catastrophic forgetting} as it primarily focuses on constraining classifiers of the old categories. 
However, excessively constraining parameter optimization can impede learning for novel categories. 
Taking all these factors into account, we designed a Background Feature Distillation~(BFD) method that only acts on the features out of the regions of current categories $\mathcal{R}_{C^t}$:
\begin{equation}
    \mathcal{L}_{BFD} = \sum_{i=1}^{t-1} \|\psi^t_b \cdot F^t_{m^i} - \psi^t_b \cdot F^{t-1}_{m^i}\|_{2}^2,  
\end{equation}
where $F^t_{m^i}$ denotes the intermediate features in $f^t_{C^i}$, $\psi^t_b$ represents the binary mask out of $\mathcal{R}_{C^t}$, and $\|\cdot\|_2$ calculates the Euclidean distance between the two matrix. 
The Background Feature Distillation allows other losses to supervise the region of novel categories $\mathcal{R}_{C^t}$ while restricting the negative effect outside $\mathcal{R}_{C^t}$, where old categories $C^{1:t-1}$ could appear. 
\end{itemize}

\subsection{Objective Function}

With our proposed losses above, we construct our objective function as follows: 
\begin{equation}
\begin{aligned}
    \mathcal{L} = \mathcal{L}_{PB-BCE} + \lambda_{1} \cdot \mathcal{L}_{BgA^+} + \lambda_{2} \cdot \mathcal{L}_{BgA^-}
    + \lambda_3 \cdot \mathcal{L}_{GKD} + \lambda_4 \cdot \mathcal{L}_{BFD},
\end{aligned}
\end{equation}
where $\mathcal{L}_{PB-BCE}$ supervises the learning of novel classes, $\mathcal{L}_{PB-BCE}$ and $\mathcal{L}_{BgA^-}$ enhance the effect of Background Adaptation, $\mathcal{L}_{GKD}$ and $\mathcal{L}_{BFD}$ mitigate forgetting old classes. Hyperparameters $\lambda_{1}, \lambda_{2}, \lambda_{3}, \lambda_{4}$ balance these constraints. 

\section{Experiments}

\subsection{Experimental Setup}

\textbf{Datasets.} We follow the settings of most previous work and evaluate our proposed methods on Pascal VOC 2012~\cite{pascal10, sds11} and ADE20K~\cite{ade17} datasets. 
Pascal VOC 2012 is composed of 20 different classes with 10,582 images for training and 1,449 images for validation. 
ADE20K has 150 classes that contain 100 thing classes and 50 stuff classes. 
There are 20,210 images for training and 2,000 images for validation in ADE20K. 

\noindent\textbf{Protocols.} 
Earlier work~\cite{mib20, plop21} has defined two types of training settings, \ie \textit{overlapping} and \textit{disjoint}. 
The \textit{overlapping} setting contains future classes in the previous steps, whereas the \textit{disjoint} setting forbids future classes until they belong to the training classes. 
We follow recent work~\cite{dkd22, alife22, idec23, ewf23} and adopt the \textit{overlapping} setting for evaluating as it is more realistic. 
Besides, following the previous training protocols~\cite{ewf23, mib20, plop21}, we set each sub-task from the dataset as $N_{ini} - N_{inc}$, where $N_{ini}$ and $N_{inc}$ denotes the quantity of the categories in the initial step and a single incremental step, respectively. 
For example, the 15-1 sub-task of Pascal VOC 2012 has 15 classes for the initial training step. 
During each incremental step, one novel class is applied for incremental training without previous classes. 
Specifically, we define 4 sub-tasks on Pascal VOC 2012, including 19-1~(2 steps), 15-1~(6 steps), 10-1~(11 steps), and 5-3~(6 steps). 
On ADE20K, we define 4 sub-tasks containing 100-50~(2 steps), 100-10~(6 steps), 50-50~(3 steps), and 100-5~(11 steps). 

\noindent\textbf{Evaluation metrics.} 
We evaluate our proposed method with widely-used mean Intersection-over-Union~(mIoU). 
The Intersection-over-Union~(IoU) is defined as $IoU = \frac{TP}{TP + FP + FN}$, 
where $TP$, $FP$, and $FN$ represent the quantity of true-positive, false-positive, and false-negative prediction pixels. 
The mIoU calculates the mean value of IoU corresponding to each category. 
To effectively measure the performance of CISS tasks, we compute the mIoU values of the initial classes $C^0$, the incremental classes $C^{1:t}$, and all classes $C^{0:t}$, respectively. 

\noindent\textbf{Implementation details.} 
Following the previous studies~\cite{mib20, sdr21, ssul21}, we apply Deeplab-v3~\cite{deeplabv3} as the segmentation network with a ResNet-101~\cite{resnet16} for feature extraction. 
The ResNet-101 is initially pretrained on ImageNet-1K~\cite{imagenet}. 
We use SGD optimizer to optimize the network. 
The initial learning rate of the initial step is set to $10^{-2}$, while in the incremental steps, the initial learning rate is set to $10^{-3}$ for Pascal VOC 2012 and $10^{-2}$ for ADE20K. 
The momentum value is 0.9 and the weight decay is $10^{-4}$ in all steps. 
We apply the poly learning rate schedule following~\cite{deeplabv3, dkd22, ssul21, plop21}. 
During the initial step, the training epoch is set to 50 for Pascal VOC 2012 and 60 for ADE20K, while for each incremental step, we train the network for 20 epochs on Pascal VOC and 100 epochs on ADE20K.  
The batch size is 16 for Pascal VOC 2012 and 8 for ADE20K in all steps. 
The data augmentation follows the standard of previous work, which is composed of random scaling, random flipping, and random crop. 
The coefficients $\lambda_1$, $\lambda_2$, $\lambda_3$, $\lambda_4$, of the objective function are set to 1, 5, 1, 4. 
The threshold of pseudo labeling $\tau=0.7$ is set to all settings.
The implementation is built on Pytorch. The experiments are conducted on NVIDIA RTX2080Ti GPUs. 

\subsection{Quantitative Results}

\noindent\textbf{Comparison on Pascal VOC 2012.} 
We evaluate our proposed method with current exemplar-free DeeplabV3-based methods on the Pascal VOC 2012 dataset. The sub-tasks of Pascal VOC 2012 have distinct characteristics. 
For example, the sub-task 19-1 has more categories in the initial step, while 5-3 has fewer classes for the initial training and more incremental steps, which is much more challenging. 
The results in Tab.~\ref{tab:voc} show that our proposed method outperforms other methods on novel classes and overall mIoU results of all sub-tasks. 
Particularly, our approach has $1.7\%$ and $3.0\%$ promotion of mIoU in more challenging sub-tasks of 10-1 and 5-3, respectively. 
Except for DKD~\cite{dkd22} which refine the background class, our approach achieves at least 3.0\% and 7.0\% of promotion on mIoU Compared to the methods with a standard shared background classifier, demonstrating the effectiveness of the Background Adaptation mechanism. 
Besides, our method achieve an impressive advantage on the novel classes, with significant improvements of 4.5\% and 3.5\% compared to the previous state-of-the-art method. 
Many previous methods emphasize restricting catastrophic forgetting by boosting the mIoU of larger-proportion old classes in specific tasks (\eg 15-1) to make the overall mIoU appear to be higher. 
However, our approach not only retains the ability to segment old classes but also significantly improves the accuracy of novel classes, aligning more closely with the \textit{incremental} objective emphasized in CISS. 

\begin{table}[!t]
    \centering
    \caption{Quantitative comparison with state-of-the-art exemplar-free methods on Pascal VOC 2012 in mIoU. 
    Scores of novel classes and all classes in \textbf{bold} are the best while \underline{underlined} are the second best. }
    \begin{tabular}{r|cc|c|cc|c|cc|c|cc|c}
    \toprule
         \multirow{2}{*}{Method}&  \multicolumn{3}{c|}{\textbf{15-1} (6 steps)}&  \multicolumn{3}{c|}{\textbf{10-1} (11 steps)}&  \multicolumn{3}{c|}{\textbf{5-3} (6 steps)}&  \multicolumn{3}{c}{\textbf{19-1} (1 step)}\\
         &  0-15&  16-20&  all&  0-10&  11-20&  all&  0-5&  6-20&  all&  0-19&  20& all\\
         \midrule
         MiB~\pub{CVPR20}~\cite{mib20}& 38.0 & 13.5 & 32.2 &  12.2&  13.1&  12.6&  57.1&  42.5&  46.7&  71.2&  22.1& 68.9\\
         SDR~\pub{CVPR21}~\cite{sdr21} &  47.3&  14.7&  39.5&  32.4&  17.1&  25.1& - & - & - &  69.1&  32.6& 67.4\\
         PLOP~\pub{CVPR21}~\cite{plop21} &  65.1&  21.1&  54.6&  44.0&  15.5&  30.5&  25.7&  30.0&  28.7&  75.4&  37.3& 73.5\\
         RCIL~\pub{CVPR22}~\cite{rcil22} &  70.6&  23.7&  59.4&  55.4&  15.1&  34.3&  63.1&  34.6&  42.8&  77.0&  31.5& 74.7\\
         EWF~\pub{CVPR23}~\cite{ewf23} &  77.7&  32.7&  67.0&  71.5&  30.3&  51.9&  61.7&  42.2&  47.7&  77.9&  6.7& 74.5\\
         AWT~\pub{WACV23}~\cite{awt23} &  59.1&  17.2&  49.1&  33.2&  18.0&  26.0&  61.8&  45.9&  50.4& - & - & - \\         IDEC~\pub{TPAMI23}~\cite{idec23} &  77.0&  36.5&  67.3&  70.7&  46.3&  59.1&  67.1&  49.0&  54.1& - & - & -\\         DKD~\pub{NeurIPS22}~\cite{dkd22} &  78.1&  \underline{42.7}&  \underline{69.7}&  73.1&  \underline{46.5}&  \underline{60.4}&  69.6&  \underline{53.5}&  \underline{58.1} &  77.8&  \underline{41.5}& \underline{76.0}\\
         \hline
         Ours &  77.6&  \textbf{45.9}&  \textbf{70.0}&  72.2&  \textbf{51.0}&  \textbf{62.1}&  71.3&  \textbf{57.0}&  \textbf{61.1}&  78.2&  \textbf{42.2}& \textbf{76.4}\\
         \hline
 \joint{Joint}& \joint{79.8}& \joint{72.4}& \joint{77.4}& \joint{78.4}& \joint{76.4}& \joint{77.4}& \joint{76.9}& \joint{77.6}& \joint{77.4}& \joint{77.5}& \joint{77.0}&\joint{77.4}\\
 \bottomrule
    \end{tabular}
    \label{tab:voc}
\end{table}

\begin{figure}[!t]
    \centering
        \subcaptionbox{Task 15-1 of VOC2012. \label{fig:151}}{		\includegraphics[width=0.42\linewidth]{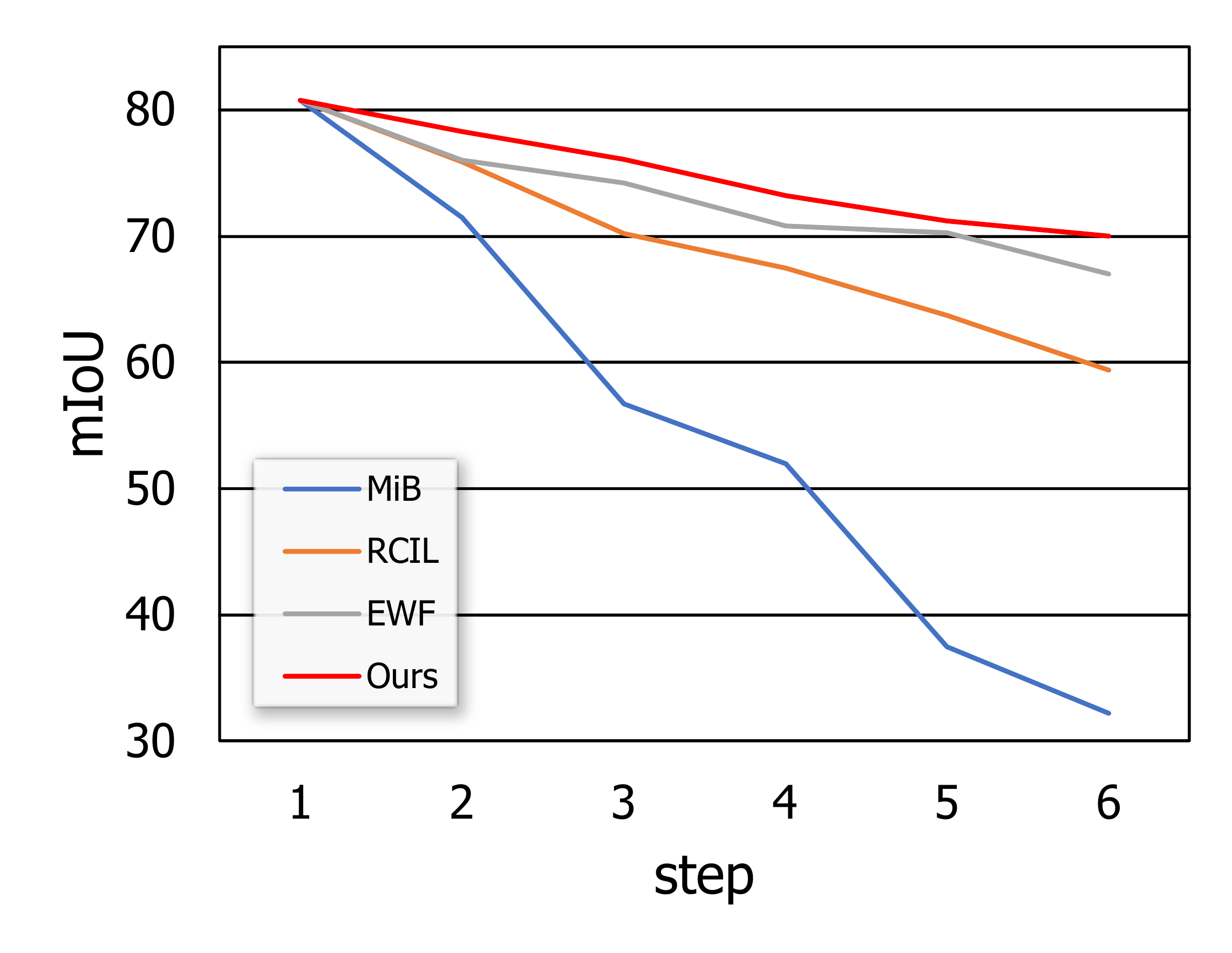}
	}
        \hspace{0.1cm}
	\subcaptionbox{Task 10-1 of VOC2012. \label{fig:101}}{		\includegraphics[width=0.44\linewidth]{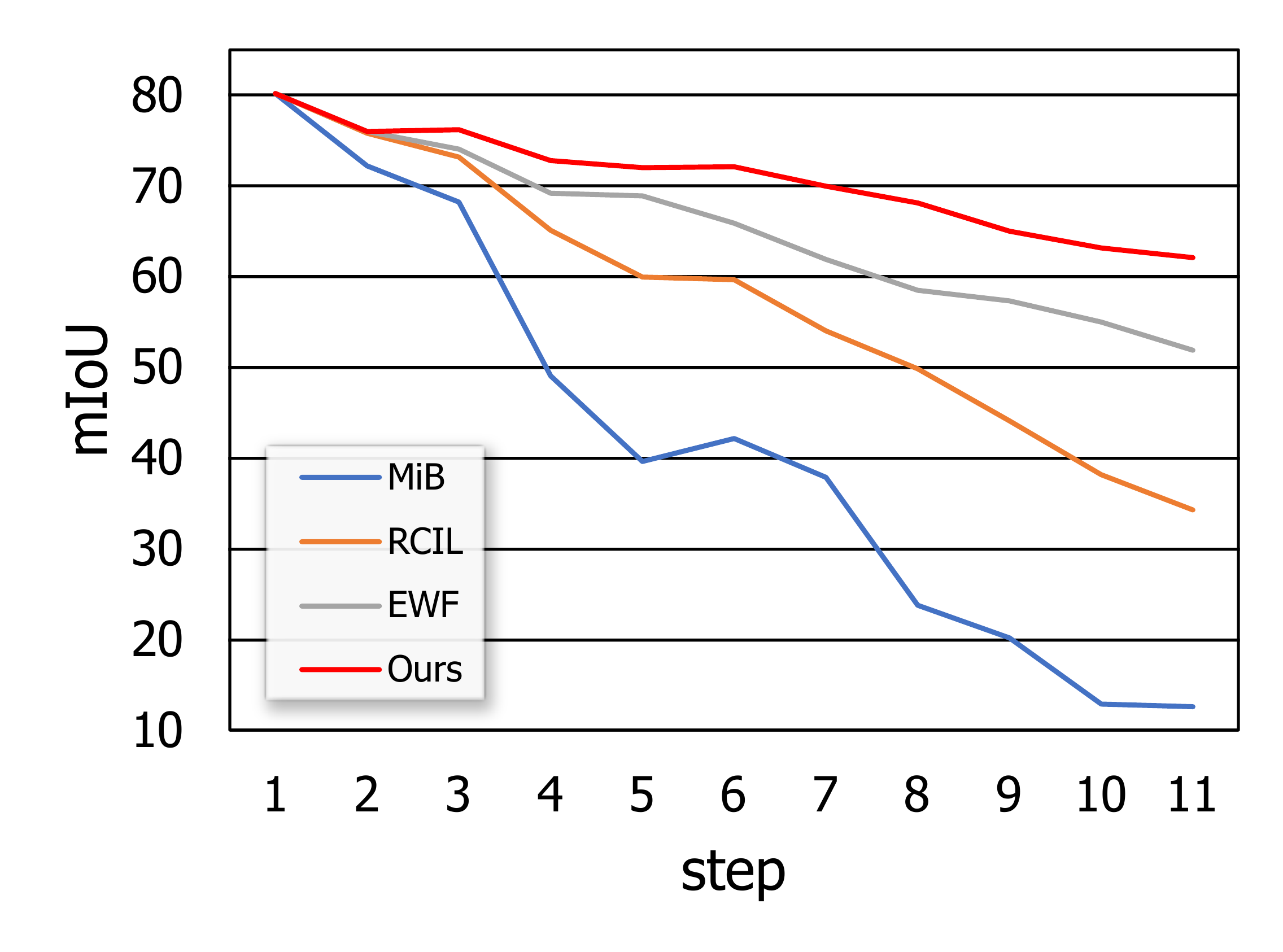}	
	}
    \caption{The step-wise mIoU comparison of our approach with previous methods under the sub-tasks 10-1 and 15-1 of Pascal VOC 2012. }
    \label{fig:steps}
\end{figure}

\begin{table}[!t]
    \centering
    \caption{Quantitative comparison with state-of-the-art exemplar-free methods on ADE20K in mIoU. 
    Scores of novel classes and all classes in \textbf{bold} are the best while \underline{underlined} are the second best. }
    \begin{tabular}{r|cc|c|cc|c|cc|c|cc|c}
    \toprule
         \multirow{2}{*}{Method}&  \multicolumn{3}{c|}{\textbf{100-50} (2 steps)}&  \multicolumn{3}{c|}{\textbf{100-10} (6 steps)}&  \multicolumn{3}{c|}{\textbf{50-50} (3 steps)}&  \multicolumn{3}{c}{\textbf{100-5} (11 steps)}\\
         &  0-100&  101-150&  all&  0-100&  101-150&  all&  0-50&  51-150&  all&  0-100&  101-150& all\\
         \midrule
         MiB~\cite{mib20}& 40.5& 17.2& 32.8&  38.2&  11.1&  29.2&  45.6&  21.0&  29.3&  36.0&  5.7& 26.0\\
         PLOP~\cite{plop21} &  41.9&  14.9&  32.9&  40.5&  13.6&  31.6&  48.8&  21.0&  30.4&  39.1&  7.8& 28.8\\
         RCIL~\cite{rcil22} &  42.3&  18.8&  34.5&  39.3&  17.6&  32.1&  48.3&  25.0&  32.5&  38.5&  11.5& 29.6\\
         EWF~\cite{ewf23} &  41.2&  21.3&  34.6&  41.5&  16.6&  33.2& - & - & - &  41.4&  13.4& \underline{32.1}\\
         AWT~\cite{awt23} &  40.9&  24.7&  35.6&  39.1&  \underline{21.4}&  33.2&  46.6&  \textbf{27.0}&  33.5&  38.6&  \underline{16.0}& 31.1\\
         GSC~\cite{gsc23} &  42.4&  19.2&  34.8&  40.8&  16.2&  32.6&  46.2&  26.4&  33.0& - & - & -\\
         IDEC~\cite{idec23} &  42.0&  18.2&  34.1&  40.3&  17.6&  32.7&  47.4&  26.0&  33.1&  39.2&  14.6& 31.0\\
         DKD~\cite{dkd22} &  42.4&  \underline{22.9}&  \textbf{36.0}&  41.6&  19.5&  \underline{34.3}&  48.8&  26.3&  \textbf{33.9}& - & - & -\\
         \hline
         Ours & 42.0 & \textbf{23.0} & \underline{35.7} & 41.1 & \textbf{23.1} & \textbf{35.2} & 47.9 & \underline{26.5}& \underline{33.7}& 40.5 & \textbf{21.2}& \textbf{34.1} \\
         \hline
 \joint{Joint}& \joint{44.3}& \joint{28.2}& \joint{38.9}& \joint{44.3}& \joint{28.2}& \joint{38.9}& \joint{51.1}& \joint{33.3}& \joint{38.9}& \joint{44.3}& \joint{28.2}&\joint{38.9}\\
 \bottomrule
    \end{tabular}
    \label{tab:ade}
\end{table}

\noindent\textbf{Comparison on ADE20K.} We evaluate our method on the sub-tasks of more challenging ADE20K, including 100-50, 100-10, 100-5, and 50-50, as shown in Tab. \ref{tab:ade}. 
Although our method is slightly lower than the state-of-the-art method in short-term sub-tasks, we are remarkably ahead of other methods in more challenging long-term 100-10 and 100-5 tasks. 
The performance in the novel classes is still outstanding, which demonstrates the universality of our method.

\subsection{Ablation Studies}

\subsubsection{Component Ablations.}\label{sec:com}

In this section, we analyze our proposed components, including the Background Adaptation mechanism, Background Feature Distillation loss $\mathcal{L}_{BFD}$, Group Knowledge Distillation loss $\mathcal{L}_{GKD}$, Background Adaptation losses $\mathcal{L}_{BgA^-}$ and $\mathcal{L}_{BgA^+}$. 
The experiments are carried out on the representative sub-task 15-1 of Pascal VOC 2012. 
We separately evaluate different combinations of components and present the results in Tab.~\ref{tab:ab_com}. 
The baseline method applies pseudo-labeling, $\mathcal{L}_{PB-BCE}$, and freezes the whole network except the classifiers of background and novel classes. 
Under the same experimental settings, the Background Adaptation mechanism significantly promotes 5.2\% of overall mIoU, demonstrating its effectiveness in retaining the knowledge of old classes, and more notably, improving the performance of novel classes. 
The usage of distillation strategies with $\mathcal{L}_{BFD}$ and $\mathcal{L}_{GKD}$ preserves the performance of old classes while further boosting mIoU by 6.0\% on novel classes. 
Moreover, the precise supervision of the Background Adaptation using $\mathcal{L}_{BgA^-}$ and $\mathcal{L}_{BgA^+}$ once again enhances the segmentation of novel classes with an additional 9.4\% of mIoU. 
Subsequently, there is a growth of 2.4\% on the overall mIoU, which proves that additional precise supervision is necessary. 
The results above validate the effectiveness of the Background Adaptation mechanism and other related losses. 
There is a notable improvement, particularly in the performance of novel classes, exceeding twice that of the baseline.

\begin{table}[!t]
    \centering
    \caption{Ablation Study on the components of our proposed method on 15-1. 
    BgA denotes Background Adaptation. 
    In particular, the experiments \textit{w/o} distillation losses~(\ie $\mathcal{L}_{BFD}$ and $\mathcal{L}_{GKD}$) freeze the whole network except the classifier $f^t$ following the settings in \cite{ssul21}. 
    All experiments contain $\mathcal{L}_{PB-BCE}$. }
    \adjustbox{width=0.6\linewidth}{
    \begin{tabular}{ccccc|ccc}
        \toprule
        \small
         BgA&  $\mathcal{L}_{BFD}$& $\mathcal{L}_{GKD}$ & $\mathcal{L}_{BgA^-}$ & $\mathcal{L}_{BgA^+}$ & mIoU\textsubscript{ini}& mIoU\textsubscript{inc} & mIoU \\
         \midrule
         &  &  &  &  &  73.4&  21.8& 61.1\\
         \checkmark &  &  &  &  &  77.4&  30.5& 66.3\\
         \checkmark &  \checkmark&  &  &  &  76.7&  37.1& 67.3\\
         \checkmark &  \checkmark&  \checkmark&  &  &  77.3&  36.5& 67.6\\
         \checkmark &  \checkmark&  \checkmark&  \checkmark&  &  77.1&  43.4& 69.0\\
         \checkmark &  \checkmark&  \checkmark&  &  \checkmark&  76.6&  34.1& 66.5\\
         
        \checkmark& \checkmark& \checkmark& \checkmark& \checkmark& 77.6& \textbf{45.9}&\textbf{70.0}\\
        \bottomrule
    \end{tabular}
    }
    \label{tab:ab_com}
\end{table}

\begin{table*}[t] 
    \centering
    \begin{minipage}{0.49\linewidth} 
        \centering
        \caption{Ablation study on schemes of Background Adaptation.}
        \adjustbox{width=0.7\linewidth}{
        \begin{tabular}{rc|ccc}
            \toprule
            & Filter & mIoU\textsubscript{ini} & mIoU\textsubscript{inc} & mIoU \\
            \midrule
            \multirow{2}{*}{MSE-0} & \xmark & 74.2 & 41.8 & 66.5 \\
                                     & \checkmark & 74.6 & 42.2 & 66.9 \\
            \hline
            BCE-1 & \checkmark & 77.4 & 44.7 & 69.6 \\
            \hline
            \multirow{2}{*}{Ours} & \xmark & 65.7 & 45.4 & 60.8 \\
                                    & \checkmark & 77.6 & \textbf{45.9} & \textbf{70.0} \\
            \bottomrule
        \end{tabular}}
        \label{tab:bga}
    \end{minipage}
    \hfill 
    \begin{minipage}{0.49\linewidth} 
        \centering
        \caption{Ablation Study on the feature distillation (FD) strategies. }
        \adjustbox{width=0.7\linewidth}{
        \begin{tabular}{c|ccc}
            \toprule
            & mIoU\textsubscript{ini} & mIoU\textsubscript{inc} & mIoU \\
            \midrule
            \textit{w/o} FD & 56.9 & 33.2 & 51.2 \\
            $\mathcal{L}_{KD}$ & 70.1 & 38.3 & 62.6 \\
            $\mathcal{L}_{MSE}$ & 77.5 & 44.3 & 69.6 \\
            $\mathcal{L}_{BFD}$ & 77.6 & \textbf{45.9} & \textbf{70.0} \\
            \bottomrule
        \end{tabular}}
        \label{tab:ofd}
    \end{minipage}
\end{table*}

\subsubsection{Background Adaptation Schemes.}

Sec.~\ref{sec:com} has presented the importance of the Background Adaptation losses. 
We have devised multiple schemes to achieve more optimal results of Background Adaptation, as shown in Tab.~\ref{tab:bga}. 
Initially, we aim to have the Background Adaptation value for the current ground truth background region equal to 0, using Mean Square Error~(MSE) loss to minimize the deviation from 0~(MSE-0). 
However, MSE-0 couldn't restrict the background value to absolute 0, leading to noise accumulation and a lower mIoU compared to Background Adaptation \textit{w/} knowledge distillation. 
Later, we incorporate considerations for the current class region and introduce a Binary Cross-Entropy~(BCE) loss~(BCE-1). 
The expectation is to have the value in the current class region as small as possible, while the value in the background class region is as large as possible. 
Besides, we apply the \textit{Filter} to remove the noise above 0. 
The BCE-1 scheme with Filter improved mIoU by $2\%$, but supervision in the background region tends to overfit to background objects. 
Therefore, we retain the BCE loss in the novel class region and redesign a triplet loss as in Sec.~\ref{sec:bgaloss}. 
The final Background Adaptation losses further enhance the segmentation capability for novel classes, resulting in a $1.2\%$ improvement.

\subsubsection{Background Feature Distillation.}

To validate the effectiveness of Background Feature Distillation loss $\mathcal{L}_{BFD}$ in Sec.~\ref{sec:kd}, we compare $\mathcal{L}_{BFD}$ with the method \textit{w/o} feature distillation~(\textit{w/o} FD), \textit{w/} standard knowledge distillation~($\mathcal{L}_{KD}$), and \textit{w/} mean square error~($\mathcal{L}_{MSE}$). 
As shown in Tab. \ref{tab:ofd}, our $\mathcal{L}_{BFD}$ outperforms other representative distillation methods. 
Particularly, the Background distillation with MSE, \ie, $\mathcal{L}_{BFD}$, shows a 1.6\% promotion of mIoU on novel classes with no decline in old classes compared to $\mathcal{L}_{MSE}$. 
This demonstrates the necessity of preserving a tunnel for learning novel classes. 


\subsection{Qualitative Analysis}\label{sec:qual}

\begin{figure}[t]
    \centering
    \includegraphics[width=0.86\linewidth]{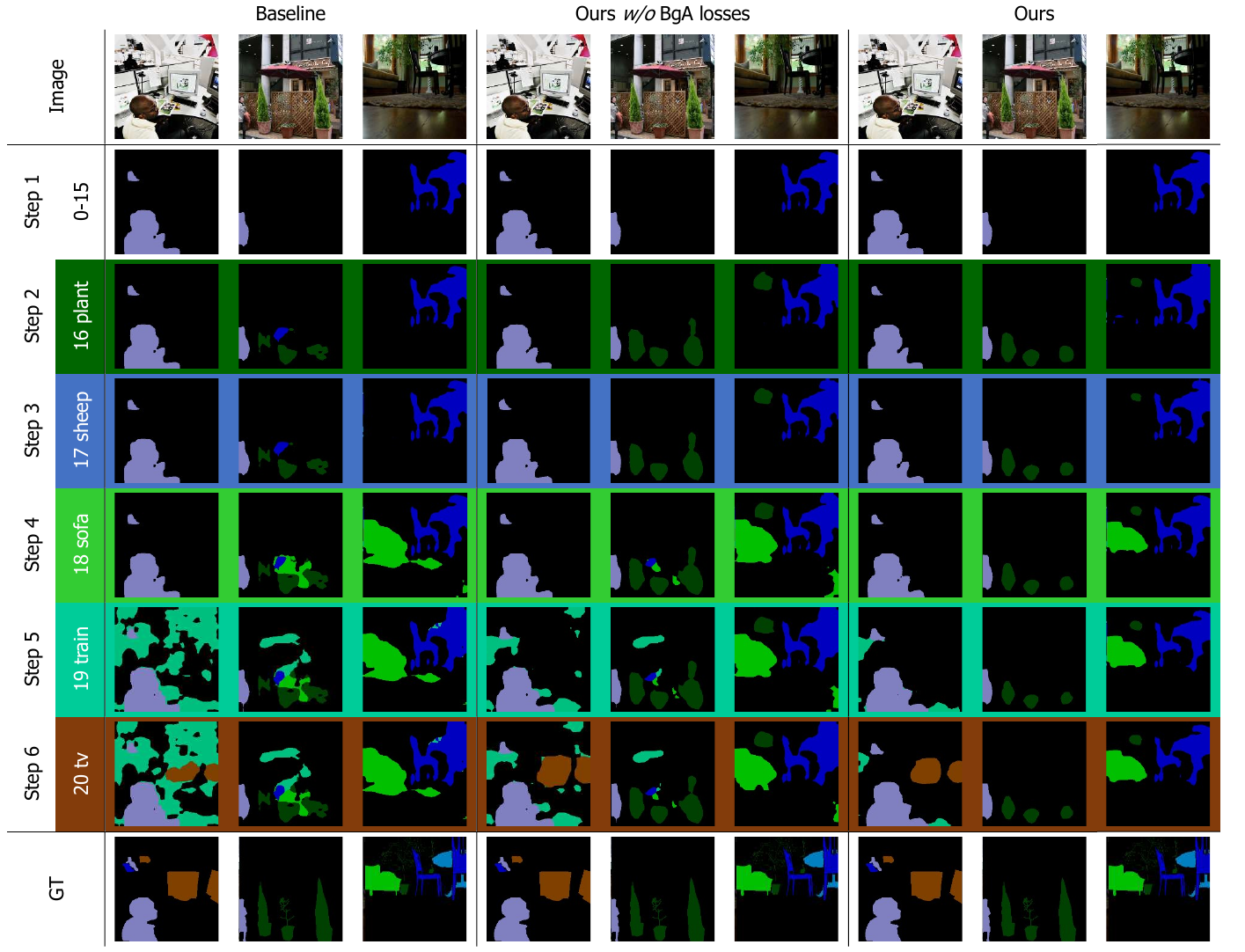}
    \caption{Qualitative analysis on 15-1 of Pascal VOC 2012. 
    The \textit{Baseline} and \textit{Ours w/o BgA losses} are in the $1^{st}$ row and the $4^{th}$ row of Tab.~\ref{tab:ab_com}. 
    $Ours$ applies additional Background Adaptation losses, compared to the \textit{Ours w/o BgA losses}.}
    \label{fig:qual}
\end{figure}

We've conducted a qualitative analysis of our approach with \textit{Baseline} and \textit{Ours w/o BgA losses} in the $1^{st}$ row and the $4^{th}$ row of Tab. \ref{tab:ab_com}. 
The results shown in Fig. \ref{fig:qual} are visualized from the 15-1 sub-task of VOC 2012. 
Our approach in the last three columns reduces the misclassification~(\eg, \textit{train} in the first two columns of each method) and enhances the precision~(\eg, \textit{sofa} in the last column of each method) of the novel classes, while maintaining segmentation ability on the old classes. 
The results showcase the stability and plasticity of our approach. 

\section{Conclusions}

We have introduced a novel CISS method featuring a Background Adaptation mechanism, designed to prevent the continual shifting of a shared background class and promote the learning of novel classes.
The mechanism explicitly models the background residual rather than the background at each step and aggregates these residuals to represent the evolving background.
To further enhance the performance of the Background Adaptation, we customized corresponding losses on the adaptation channel and improved the distillation scheme.
Experiments on our proposed components verify their effectiveness under the Background Adaptation mechanism. 
Moreover, our method surpasses previous state-of-the-art methods in most sub-tasks, notably excelling in performance of novel classes.

\clearpage  
\section*{Acknowledgement}
This work was supported by the National Natural Science Foundation of China (No. 61972036) and the Industry-University-Institute Cooperation Foundation of the Eighth Research Institute of China Aerospace Science and Technology Corporation (No. SAST2022-049).

%
%
\bibliographystyle{splncs04}
\bibliography{main}

\begin{thebibliography}{10}
\providecommand{\url}[1]{\texttt{#1}}
\providecommand{\urlprefix}{URL }
\providecommand{\doi}[1]{https://doi.org/#1}

\bibitem{ucl19}
Ahn, H., Cha, S., Lee, D., Moon, T.: Uncertainty-based continual learning with adaptive regularization. NeurIPS  \textbf{32} (2019)

\bibitem{gss19}
Aljundi, R., Lin, M., Goujaud, B., Bengio, Y.: Gradient based sample selection for online continual learning. NeurIPS  \textbf{32} (2019)

\bibitem{dkd22}
Baek, D., Oh, Y., Lee, S., Lee, J., Ham, B.: Decomposed knowledge distillation for class-incremental semantic segmentation. NeurIPS  \textbf{35},  10380--10392 (2022)

\bibitem{ccbo20}
Borsos, Z., Mutny, M., Krause, A.: Coresets via bilevel optimization for continual learning and streaming. NeurIPS  \textbf{33},  14879--14890 (2020)

\bibitem{twf22}
Boschini, M., Bonicelli, L., Porrello, A., Bellitto, G., Pennisi, M., Palazzo, S., Spampinato, C., Calderara, S.: Transfer without forgetting. In: ECCV. pp. 692--709 (2022)

\bibitem{mib20}
Cermelli, F., Mancini, M., Bulo, S.R., Ricci, E., Caputo, B.: Modeling the background for incremental learning in semantic segmentation. In: CVPR. pp. 9233--9242 (2020)

\bibitem{co2l21}
Cha, H., Lee, J., Shin, J.: Co2l: Contrastive continual learning. In: ICCV. pp. 9516--9525 (2021)

\bibitem{ssul21}
Cha, S., Yoo, Y., Moon, T., et~al.: Ssul: Semantic segmentation with unknown label for exemplar-based class-incremental learning. NeurIPS  \textbf{34},  10919--10930 (2021)

\bibitem{star23}
Chen, J., Cong, R., Luo, Y., Ip, H., Kwong, S.: Saving 100x storage: Prototype replay for reconstructing training sample distribution in class-incremental semantic segmentation. Advances in Neural Information Processing Systems  \textbf{36} (2024)

\bibitem{deeplabv1}
Chen, L.C., Papandreou, G., Kokkinos, I., Murphy, K., Yuille, A.L.: Semantic image segmentation with deep convolutional nets and fully connected crfs. arXiv preprint arXiv:1412.7062  (2014)

\bibitem{deeplabv2}
Chen, L.C., Papandreou, G., Kokkinos, I., Murphy, K., Yuille, A.L.: Deeplab: Semantic image segmentation with deep convolutional nets, atrous convolution, and fully connected crfs. IEEE TPAMI  \textbf{40}(4),  834--848 (2017)

\bibitem{deeplabv3}
Chen, L.C., Papandreou, G., Schroff, F., Adam, H.: Rethinking atrous convolution for semantic image segmentation. arXiv preprint arXiv:1706.05587  (2017)

\bibitem{chen2018encoder}
Chen, L.C., Zhu, Y., Papandreou, G., Schroff, F., Adam, H.: Encoder-decoder with atrous separable convolution for semantic image segmentation. In: ECCV. pp. 801--818 (2018)

\bibitem{lifelong18}
Chen, Z., Liu, B.: Lifelong machine learning, vol.~1. Springer (2018)

\bibitem{mask2former}
Cheng, B., Misra, I., Schwing, A.G., Kirillov, A., Girdhar, R.: Masked-attention mask transformer for universal image segmentation. In: CVPR. pp. 1290--1299 (2022)

\bibitem{maskformer}
Cheng, B., Schwing, A., Kirillov, A.: Per-pixel classification is not all you need for semantic segmentation. NeurIPS  \textbf{34},  17864--17875 (2021)

\bibitem{gsc23}
Cong, W., Cong, Y., Dong, J., Sun, G., Ding, H.: Gradient-semantic compensation for incremental semantic segmentation. arXiv preprint arXiv:2307.10822  (2023)

\bibitem{imagenet}
Deng, J., Dong, W., Socher, R., Li, L.J., Li, K., Fei-Fei, L.: Imagenet: A large-scale hierarchical image database. In: CVPR. pp. 248--255 (2009)

\bibitem{lwm19}
Dhar, P., Singh, R.V., Peng, K.C., Wu, Z., Chellappa, R.: Learning without memorizing. In: CVPR. pp. 5138--5146 (2019)

\bibitem{vit20}
Dosovitskiy, A., Beyer, L., Kolesnikov, A., Weissenborn, D., et~al.: An image is worth 16x16 words: Transformers for image recognition at scale. In: ICLR (2021)

\bibitem{plop21}
Douillard, A., Chen, Y., Dapogny, A., Cord, M.: Plop: Learning without forgetting for continual semantic segmentation. In: CVPR. pp. 4040--4050 (2021)

\bibitem{ada22}
Ermis, B., Zappella, G., Wistuba, M., Rawal, A., Archambeau, C.: Memory efficient continual learning with transformers. NeurIPS  \textbf{35},  10629--10642 (2022)

\bibitem{pascal10}
Everingham, M., Van~Gool, L., Williams, C.K., Winn, J., Zisserman, A.: The pascal visual object classes (voc) challenge. IJCV  \textbf{88},  303--338 (2010)

\bibitem{ln23}
Fang, Y., Zhu, F., Cheng, B., Liu, L., Zhao, Y., Wei, Y.: Locating noise is halfway denoising for semi-supervised segmentation. In: Proceedings of the IEEE/CVF International Conference on Computer Vision. pp. 16612--16622 (2023)

\bibitem{awt23}
Goswami, D., Schuster, R., van~de Weijer, J., Stricker, D.: Attribution-aware weight transfer: A warm-start initialization for class-incremental semantic segmentation. In: Proceedings of the IEEE/CVF Winter Conference on Applications of Computer Vision. pp. 3195--3204 (2023)

\bibitem{sds11}
Hariharan, B., Arbel{\'a}ez, P., Bourdev, L., Maji, S., Malik, J.: Semantic contours from inverse detectors. In: ICCV. pp. 991--998 (2011)

\bibitem{resnet16}
He, K., Zhang, X., Ren, S., Sun, J.: Deep residual learning for image recognition. In: CVPR. pp. 770--778 (2016)

\bibitem{lfl16}
Jung, H., Ju, J., Jung, M., Kim, J.: Less-forgetting learning in deep neural networks. arXiv preprint arXiv:1607.00122  (2016)

\bibitem{ewc17}
Kirkpatrick, J., Pascanu, R., Rabinowitz, N., Veness, J., et~al.: Overcoming catastrophic forgetting in neural networks. Proceedings of the National Academy of Sciences  \textbf{114}(13),  3521--3526 (2017)

\bibitem{alexnet12}
Krizhevsky, A., Sutskever, I., Hinton, G.E.: Imagenet classification with deep convolutional neural networks. NeurIPS  \textbf{25} (2012)

\bibitem{lenet98}
LeCun, Y., Bottou, L., Bengio, Y., Haffner, P.: Gradient-based learning applied to document recognition. Proceedings of the IEEE  \textbf{86}(11),  2278--2324 (1998)

\bibitem{lwf17}
Li, Z., Hoiem, D.: Learning without forgetting. IEEE TPAMI  \textbf{40}(12),  2935--2947 (2017)

\bibitem{swin21}
Liu, Z., Lin, Y., Cao, Y., Hu, H., Wei, Y., Zhang, Z., Lin, S., Guo, B.: Swin transformer: Hierarchical vision transformer using shifted windows. In: ICCV. pp. 10012--10022 (2021)

\bibitem{fcn}
Long, J., Shelhamer, E., Darrell, T.: Fully convolutional networks for semantic segmentation. In: CVPR. pp. 3431--3440 (2015)

\bibitem{gradient17}
Lopez-Paz, D., Ranzato, M.: Gradient episodic memory for continual learning. NeurIPS  \textbf{30} (2017)

\bibitem{lump21}
Madaan, D., Yoon, J., Li, Y., Liu, Y., Hwang, S.J.: Representational continuity for unsupervised continual learning. In: ICLR (2022)

\bibitem{recall21}
Maracani, A., Michieli, U., Toldo, M., Zanuttigh, P.: Recall: Replay-based continual learning in semantic segmentation. In: ICCV. pp. 7026--7035 (2021)

\bibitem{there95}
McClelland, J.L., McNaughton, B.L., O'Reilly, R.C.: Why there are complementary learning systems in the hippocampus and neocortex: insights from the successes and failures of connectionist models of learning and memory. Psychological review  \textbf{102}(3), ~419 (1995)

\bibitem{catastrophic89}
McCloskey, M., Cohen, N.J.: Catastrophic interference in connectionist networks: The sequential learning problem. In: Psychology of learning and motivation, vol.~24, pp. 109--165. Elsevier (1989)

\bibitem{sdr21}
Michieli, U., Zanuttigh, P.: Continual semantic segmentation via repulsion-attraction of sparse and disentangled latent representations. In: CVPR. pp. 1114--1124 (2021)

\bibitem{alife22}
Oh, Y., Baek, D., Ham, B.: Alife: Adaptive logit regularizer and feature replay for incremental semantic segmentation. NeurIPS  \textbf{35},  14516--14528 (2022)

\bibitem{dgm19}
Ostapenko, O., Puscas, M., Klein, T., Jahnichen, P., Nabi, M.: Learning to remember: A synaptic plasticity driven framework for continual learning. In: CVPR. pp. 11321--11329 (2019)

\bibitem{dualnet21}
Pham, Q., Liu, C., Hoi, S.: Dualnet: Continual learning, fast and slow. NeurIPS  \textbf{34},  16131--16144 (2021)

\bibitem{icarl17}
Rebuffi, S.A., Kolesnikov, A., Sperl, G., Lampert, C.H.: icarl: Incremental classifier and representation learning. In: CVPR. pp. 2001--2010 (2017)

\bibitem{ritter2018online}
Ritter, H., Botev, A., Barber, D.: Online structured laplace approximations for overcoming catastrophic forgetting. NeurIPS  \textbf{31} (2018)

\bibitem{unet}
Ronneberger, O., Fischer, P., Brox, T.: U-net: Convolutional networks for biomedical image segmentation. In: {Proc. International Conference on Medical Image Computing and Computer-Assisted Intervention}. pp. 234--241 (2015)

\bibitem{schwarz2018progress}
Schwarz, J., Czarnecki, W., Luketina, J., Grabska-Barwinska, A., Teh, Y.W., Pascanu, R., Hadsell, R.: Progress \& compress: A scalable framework for continual learning. In: Proc. International Conference on Machine Learning. pp. 4528--4537. PMLR (2018)

\bibitem{incre23}
Shang, C., Li, H., Meng, F., Wu, Q., Qiu, H., Wang, L.: Incrementer: Transformer for class-incremental semantic segmentation with knowledge distillation focusing on old class. In: Proceedings of the IEEE/CVF Conference on Computer Vision and Pattern Recognition. pp. 7214--7224 (2023)

\bibitem{dgr17}
Shin, H., Lee, J.K., Kim, J., Kim, J.: Continual learning with deep generative replay. NeurIPS  \textbf{30} (2017)

\bibitem{dlcft22}
Shon, H., Lee, J., Kim, S.H., Kim, J.: Dlcft: Deep linear continual fine-tuning for general incremental learning. In: ECCV. pp. 513--529 (2022)

\bibitem{vggnet15}
Simonyan, K., Zisserman, A.: Very deep convolutional networks for large-scale image recognition. In: ICLR (2015)

\bibitem{segmenter21}
Strudel, R., Garcia, R., Laptev, I., Schmid, C.: Segmenter: Transformer for semantic segmentation. In: Proceedings of the IEEE/CVF international conference on computer vision. pp. 7262--7272 (2021)

\bibitem{deit21}
Touvron, H., Cord, M., Douze, M., Massa, F., Sablayrolles, A., J{\'e}gou, H.: Training data-efficient image transformers \& distillation through attention. In: Proc. International Conference on Machine Learning. pp. 10347--10357 (2021)

\bibitem{vaswani2017attention}
Vaswani, A., Shazeer, N., Parmar, N., Uszkoreit, J., Jones, L., Gomez, A.N., Kaiser, {\L}., Polosukhin, I.: Attention is all you need. NeurIPS  \textbf{30} (2017)

\bibitem{mergan18}
Wu, C., Herranz, L., Liu, X., Van De~Weijer, J., Raducanu, B., et~al.: Memory replay gans: Learning to generate new categories without forgetting. NeurIPS  \textbf{31} (2018)

\bibitem{ewf23}
Xiao, J.W., Zhang, C.B., Feng, J., Liu, X., van~de Weijer, J., Cheng, M.M.: Endpoints weight fusion for class incremental semantic segmentation. In: CVPR. pp. 7204--7213 (2023)

\bibitem{ucd22}
Yang, G., Fini, E., Xu, D., Rota, P., Ding, M., Nabi, M., Alameda-Pineda, X., Ricci, E.: Uncertainty-aware contrastive distillation for incremental semantic segmentation. IEEE TPAMI  \textbf{45}(2),  2567--2581 (2022)

\bibitem{rcil22}
Zhang, C.B., Xiao, J.W., Liu, X., Chen, Y.C., Cheng, M.M.: Representation compensation networks for continual semantic segmentation. In: CVPR. pp. 7053--7064 (2022)

\bibitem{slca23}
Zhang, G., Wang, L., Kang, G., Chen, L., Wei, Y.: Slca: Slow learner with classifier alignment for continual learning on a pre-trained model. In: Proceedings of the IEEE/CVF International Conference on Computer Vision. pp. 19148--19158 (2023)

\bibitem{side20}
Zhang, J.O., Sax, A., Zamir, A., Guibas, L., Malik, J.: Side-tuning: a baseline for network adaptation via additive side networks. In: ECCV. pp. 698--714 (2020)

\bibitem{micro22}
Zhang, Z., Gao, G., Fang, Z., Jiao, J., Wei, Y.: Mining unseen classes via regional objectness: A simple baseline for incremental segmentation. NeurIPS  \textbf{35},  24340--24353 (2022)

\bibitem{coinseg23}
Zhang, Z., Gao, G., Jiao, J., Liu, C.H., Wei, Y.: Coinseg: Contrast inter-and intra-class representations for incremental segmentation. In: Proceedings of the IEEE/CVF International Conference on Computer Vision. pp. 843--853 (2023)

\bibitem{idec23}
Zhao, D., Yuan, B., Shi, Z.: Inherit with distillation and evolve with contrast: Exploring class incremental semantic segmentation without exemplar memory. IEEE TPAMI  (2023)

\bibitem{ade17}
Zhou, B., Zhao, H., Puig, X., Fidler, S., Barriuso, A., Torralba, A.: Scene parsing through ade20k dataset. In: CVPR. pp. 633--641 (2017)

\bibitem{ctp23}
Zhu, H., Wei, Y., Liang, X., Zhang, C., Zhao, Y.: Ctp: Towards vision-language continual pretraining via compatible momentum contrast and topology preservation. In: Proceedings of the IEEE/CVF International Conference on Computer Vision. pp. 22257--22267 (2023)

\bibitem{amss23}
Zhu, L., Chen, T., Yin, J., See, S., Liu, J.: Continual semantic segmentation with automatic memory sample selection. In: CVPR. pp. 3082--3092 (2023)

\end{thebibliography}
\end{document}